\documentclass{llncs}
\usepackage{graphicx}
\usepackage{amsmath,amssymb} % define this before the line numbering.
\usepackage{color}
\usepackage[width=122mm,left=12mm,paperwidth=146mm,height=193mm,top=12mm,paperheight=217mm]{geometry}
\usepackage{enumitem}

\DeclareMathOperator{\softmax}{softmax}

\begin{document}
% \linenumbers
\pagestyle{headings}
\mainmatter

\title{Ask, Attend and Answer: Exploring Question-Guided Spatial Attention for Visual Question Answering} % Replace with your title

%\titlerunning{Ask, Attend and Answer: Exploring Question-Guided Spatial Attention for Visual Question Answering}
%\authorrunning{Huijuan Xu and Kate Saenko}

\author{Huijuan Xu \and Kate Saenko}
\institute{Department of Computer Science, UMass Lowell, USA\\{\tt\small hxu1@cs.uml.edu}, {\tt\small saenko@cs.uml.edu}}
%{\tt\small hxu1@cs.uml.edu} \and {\tt\small saenko@cs.uml.edu}

\maketitle

%%%%%%%%% ABSTRACT
\begin{abstract}
We address the problem of Visual Question Answering (VQA), which requires joint image and language understanding to answer a question about a given photograph. Recent approaches have applied deep image captioning methods based on convolutional-recurrent networks to this problem, but have failed to model spatial inference. To remedy this, we propose a model we call the Spatial Memory Network and apply it to the VQA task. Memory networks are recurrent neural networks with an explicit attention mechanism that selects certain parts of the information stored in memory. Our Spatial Memory Network stores neuron activations from different spatial regions of the image in its memory, and uses the question to choose relevant regions for computing the answer, 
a process of which constitutes a single ``hop'' in the network. 
We propose a novel spatial attention architecture that aligns words with image patches in the first hop, and obtain improved results by adding a second attention hop which considers the whole question to choose visual evidence based on the results of the first hop. 
To better understand the inference process learned by the network, we design synthetic questions that specifically require spatial inference and visualize the attention weights. We evaluate our model on two published visual question answering datasets, DAQUAR~\cite{DBLP:journals/corr/MalinowskiF14} and VQA~\cite{DBLP:journals/corr/AntolALMBZP15}, and obtain improved results compared to a strong deep baseline model (iBOWIMG) which concatenates image and question features to predict the answer~\cite{zhou2015simple}.
\keywords{Visual Question Answering, Spatial Attention, Memory Network, Deep Learning}
\end{abstract}

%%%%%%%%% BODY TEXT
\section{Introduction}
Visual Question Answering (VQA) is an emerging interdisciplinary research problem at the intersection of computer vision, natural language processing and artificial intelligence. It has many real-life applications, such as automatic querying of surveillance video~\cite{tu2014joint} or assisting the visually impaired~\cite{lasecki2014increasing}. Compared to the recently popular image captioning task~\cite{donahue2014long,vinyals2014show,karpathy2014deep,fang2014captions}, VQA requires a deeper understanding of the image, but is considerably easier to evaluate. It also puts more focus on artificial intelligence, namely the inference process needed to produce the answer to the visual question.

\begin{figure}[!t]
\centering
\includegraphics[width=0.7\columnwidth,height=2.6cm]{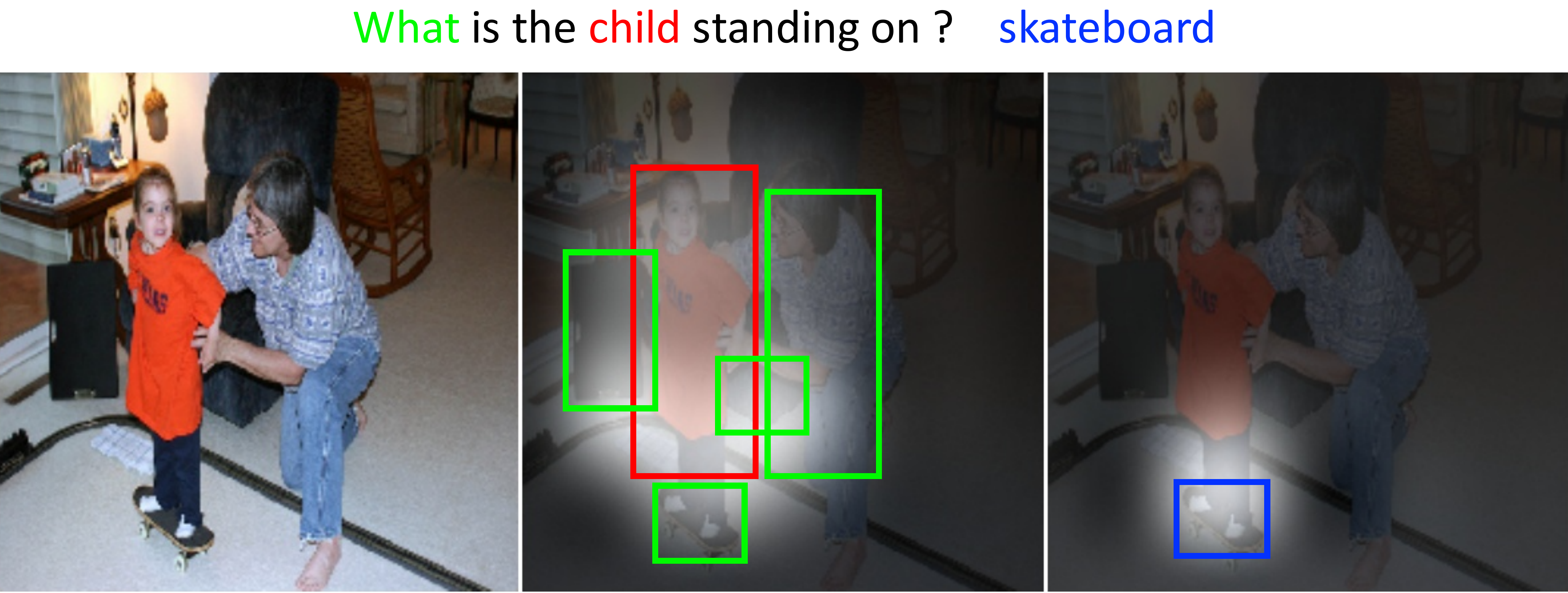}
\includegraphics[width=0.7\columnwidth,height=2.6cm]{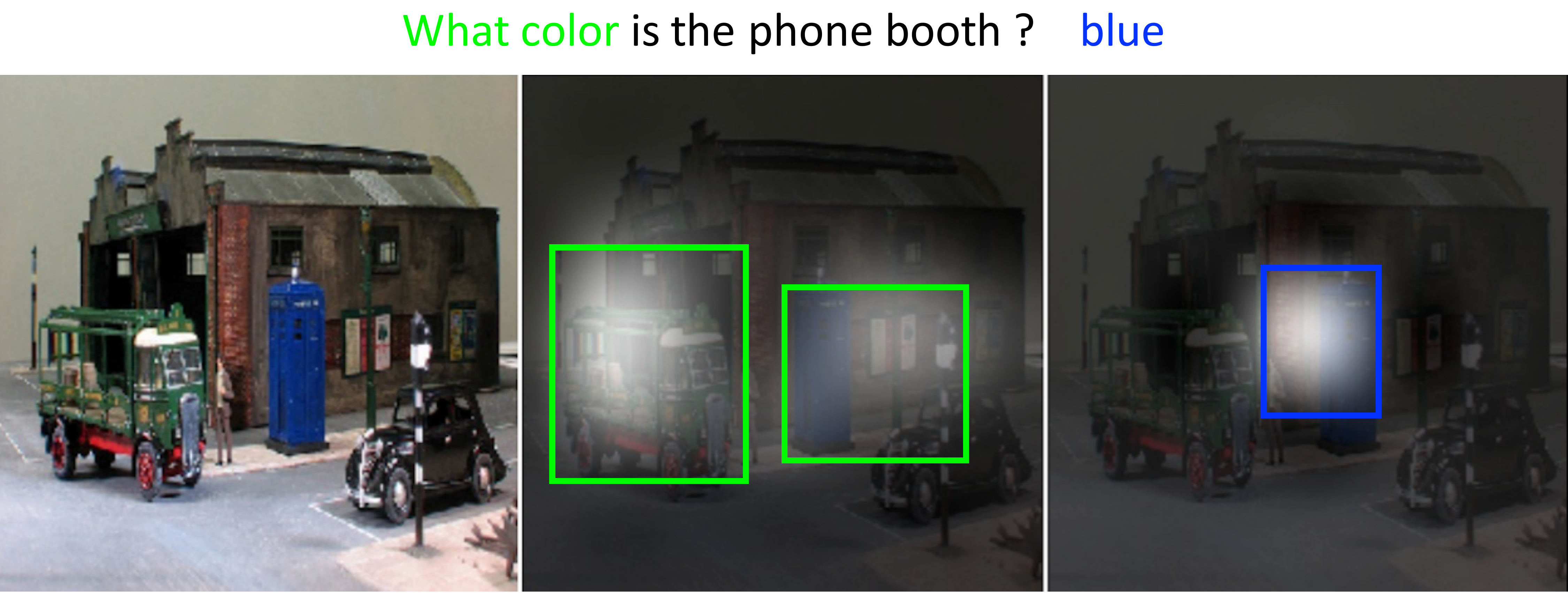}
\vspace{-0.1in}
\caption{We propose a Spatial Memory Network for VQA (SMem-VQA) that answers questions about images using spatial inference.
The figure shows the inference process of our two-hop model on examples from the VQA dataset~\cite{DBLP:journals/corr/AntolALMBZP15}. In the first hop (middle), the attention process captures the correspondence between individual words in the question and image regions. High attention regions (bright areas) are marked with bounding boxes and the corresponding words are highlighted using the same color. In the second hop (right), the fine-grained evidence gathered in the first hop, as well as an embedding of the entire question, are used to collect more exact evidence to predict the answer. (Best viewed in color.) }
\label{fig:concept}
\vspace{-0.2in}
\end{figure}

In one of the early works~\cite{DBLP:journals/corr/MalinowskiF14}, VQA is seen as a Turing test proxy. The authors propose an approach based on handcrafted features using a semantic parse of the question and scene analysis of the image combined in a latent-world Bayesian framework. More recently, several end-to-end deep neural networks that learn features directly from data have been applied to this problem~\cite{malinowski2015ask,DBLP:journals/corr/RenKZ15}.
Most of these are directly adapted from captioning models~\cite{donahue2014long,vinyals2014show,karpathy2014deep},
and utilize a recurrent LSTM network, which takes the question and Convolutional Neural Net (CNN) image features as input, and outputs the answer. Though the deep learning methods in~\cite{malinowski2015ask,DBLP:journals/corr/RenKZ15} have shown great improvement compared to the handcrafted feature method~\cite{DBLP:journals/corr/MalinowskiF14}, they have their own drawbacks. These models based on the LSTM reading in both the question and the image features do not show a clear improvement compared to an LSTM reading in the question only~\cite{malinowski2015ask,DBLP:journals/corr/RenKZ15}.   
Furthermore, the rather complicated LSTM models obtain similar or worse accuracy to a baseline model which concatenates CNN features and a bag-of-words question embedding to predict the answer, see the IMG+BOW model in~\cite{DBLP:journals/corr/RenKZ15} and the iBOWIMG model in~\cite{zhou2015simple}.

A major drawback of existing models is that they do not have any explicit notion of object position, and do not support the computation of intermediate results based on spatial attention.
Our intuition is that answering visual questions often involves looking at different spatial regions and comparing their contents and/or locations. For example, to answer the questions in Fig.~\ref{fig:concept}, we need to look at a portion of the image, such as the child or the phone booth.
Similarly, to answer the question ``Is there a cat in the basket?'' in Fig.~\ref{fig:illus}, we can first find the basket and the cat objects, and then compare their locations.

We propose a new deep learning approach to VQA that incorporates explicit spatial attention, which we call the Spatial Memory Network VQA (SMem-VQA). 
Our approach is based on memory networks, which have recently been proposed for text Question Answering (QA)~\cite{DBLP:journals/corr/WestonCB14,sukhbaatar2015end}. Memory networks combine learned text embeddings with an attention mechanism and multi-step inference. 
The text QA memory network stores textual knowledge in its ``memory" in the form of sentences, and selects relevant sentences to infer the answer. However, in VQA, the knowledge is in the form of an image, thus the memory and the question come from different modalities.
We adapt the end-to-end memory network~\cite{sukhbaatar2015end} to solve visual question answering by storing the convolutional network outputs obtained from different receptive fields into the memory, which explicitly allows spatial attention over the image. We also propose to repeat the process of gathering evidence from attended regions, enabling the model to update the answer based on several attention steps, or ``hops''. The entire model is trained end-to-end and the evidence for the computed answer can be visualized using the attention weights. 

To summarize our contributions, in this paper we\vspace{-0.1in}
\begin{itemize}%[noitemsep]
\item propose a novel multi-hop memory network with spatial attention for the VQA task which allows one to visualize the spatial inference process used by the deep network (a CAFFE~\cite{jia2014caffe} implementation will be made available), 
\item design an attention architecture in the first hop which uses each word embedding to capture fine-grained alignment between the image and question,
\item create a series of synthetic questions that explicitly require spatial inference to analyze the working principles of the network, and show that it learns logical inference rules by visualizing the attention weights,
\item provide an extensive evaluation of several existing models and our own model on the same publicly available datasets.
\end{itemize}

Sec.~2 introduces relevant work on memory networks and attention models. Sec.~3 describes our design of the multi-hop memory network architecture for visual question answering (SMem-VQA). Sec.~4  visualizes the inference rules learned by the network for synthetic spatial questions and shows the experimental results on DAQUAR~\cite{DBLP:journals/corr/MalinowskiF14} and VQA~\cite{DBLP:journals/corr/AntolALMBZP15} datasets. Sec.~5 concludes the paper.

\vspace{-0.05in}
\section{Related work}\label{sec:related}
\vspace{-0.05in}

Before the popularity of visual question answering (VQA), text question answering (QA) had already been established as a mature research problem in the area of natural language processing. Previous QA methods include searching for the key words of the question in a search engine~\cite{yahya2012natural}; parsing the question as a knowledge base (KB) query~\cite{berant2014semantic}; or embedding the question and using a similarity measurement to find evidence for the answer~\cite{bordes2014question}. 
Recently, memory networks were proposed for solving the QA problem.  \cite{DBLP:journals/corr/WestonCB14} first introduces the memory network as a general model that consists of a memory and four components: input feature map, generalization, output feature map and response. The model is investigated in the context of question answering, where the long-term memory acts as a dynamic knowledge base and the output is a textual response. 
\cite{sukhbaatar2015end} proposes a competitive memory network model that uses less supervision, called end-to-end memory network, which has a recurrent attention model over a large external memory. 
The Neural Turing Machine (NTM)~\cite{graves2014neural} couples a neural network to external memory and interacts with it by attentional processes to infer simple algorithms such as copying, sorting, and associative recall from input and output examples. 
In this paper, we solve the VQA problem using a multimodal memory network architecture that applies a spatial attention mechanism over an input image guided by an input text question. 

%\subsection{attention mechanism papers}
The neural attention mechanism has been widely used in different areas of computer vision and natural language processing, see  for example the attention models in image captioning~\cite{xu2015show}, video description generation~\cite{yao2015describing}, machine translation~\cite{bahdanau2014neural}\cite{luong2015effective} and machine reading systems~\cite{hermann2015teaching}. 
Most methods use the soft attention mechanism first proposed in~\cite{bahdanau2014neural}, which adds a layer to the network that predicts soft weights and uses them to compute a weighted combination of the items in memory. 
The two main types of soft attention mechanisms differ in the  function that aligns the input feature vector and the candidate feature vectors in order to compute the soft attention weights. 
The first type uses an alignment function based on ``concatenation'' of the input and each candidate (we use the term ``concatenation'' as described~\cite{luong2015effective}), and the second type uses an alignment function based on the dot product of the input and each candidate. 
The ``concatenation'' alignment function adds one input vector (e.g. hidden state vector of the LSTM) to each candidate feature vector, embeds the resulting vectors into scalar values, and then applies the softmax function to generate the attention weight for each candidate. 
\cite{xu2015show}\cite{yao2015describing}\cite{bahdanau2014neural}\cite{hermann2015teaching} use the ``concatenation'' alignment function in their soft attention models and \cite{cho2015describing} gives a literature review of such models applied to different tasks. 
On the other hand, the dot product alignment function first projects both inputs to a common vector embedding space, then takes the dot product of the two input vectors, and applies a softmax function to the resulting scalar value to produce the attention weight for each candidate. 
The end-to-end memory network~\cite{sukhbaatar2015end} uses the dot product alignment function. 
In~\cite{luong2015effective}, the authors compare these two alignment functions in an attention model for the neural machine translation task, and find that their implementation of the ``concatenation'' alignment function does not yield good performance on their task. 
Motivated by this, in this paper we use the dot product alignment function in our Spatial Memory Network. 

VQA is  related to image captioning. Several early papers about VQA directly adapt the image captioning models to solve the VQA problem~\cite{malinowski2015ask}\cite{DBLP:journals/corr/RenKZ15} by generating the answer using a recurrent LSTM network conditioned on the CNN output. But these models' performance is still limited~\cite{malinowski2015ask}\cite{DBLP:journals/corr/RenKZ15}.
\cite{zhu2015visual7w} proposes a new dataset and uses a similar attention model to that in image captioning~\cite{xu2015show}, but does not give results on the more common VQA benchmark~\cite{DBLP:journals/corr/AntolALMBZP15}, and our own implementation of this model is less accurate on~\cite{DBLP:journals/corr/AntolALMBZP15} than other baseline models. 
\cite{zhou2015simple} summarizes several recent  papers reporting results on the VQA dataset~\cite{DBLP:journals/corr/AntolALMBZP15} on arxiv.org and gives a simple but strong baseline model (iBOWIMG) on this dataset. This simple baseline concatenates the image features with the bag of word embedding question representation and feeds them into a softmax classifier to predict the answer. The iBOWIMG model beats most VQA models considered in the paper. Here, we compare our proposed model to the VQA models (namely, the ACK model~\cite{wu2015ask} and the DPPnet model~\cite{noh2015image}) which have comparable or better results than the iBOWIMG model. 
The ACK model in~\cite{wu2015ask} is essentially the same as the LSTM model in~\cite{DBLP:journals/corr/RenKZ15}, except that it uses image attribute features, the generated image caption and relevant external knowledge from a knowledge base as the input to the LSTM's first time step. 
The DPPnet model in~\cite{noh2015image} tackles VQA by learning a convolutional neural network (CNN) with some parameters predicted from a separate parameter prediction network. Their parameter prediction network uses a Gate Recurrent Unit (GRU) to generate a question representation, and based on this question input, maps the predicted weights to CNN via hashing. 
Neither of these models~\cite{wu2015ask}\cite{noh2015image} contain a spatial attention mechanism, and they both use external data in addition to the VQA dataset~\cite{DBLP:journals/corr/AntolALMBZP15}, e.g. the knowledge base in~\cite{wu2015ask} and the large-scale text corpus used to pre-train the GRU question representation~\cite{noh2015image}. In this paper, we explore a complementary approach of spatial attention to both improve performance and visualize the network's inference process, and obtain improved results without using external data compared to the iBOWIMG model~\cite{zhou2015simple} as well as the ACK model~\cite{wu2015ask} and the DPPnet model~\cite{noh2015image} which use external data. 
\vspace{-0.05in}
\section{Spatial Memory Network for VQA}\label{sec:smem}
\vspace{-0.05in}

%putting overview figure here to force it to appear on first Approach page
\begin{figure*}[t!]
\centering
\includegraphics[width=.9\linewidth]{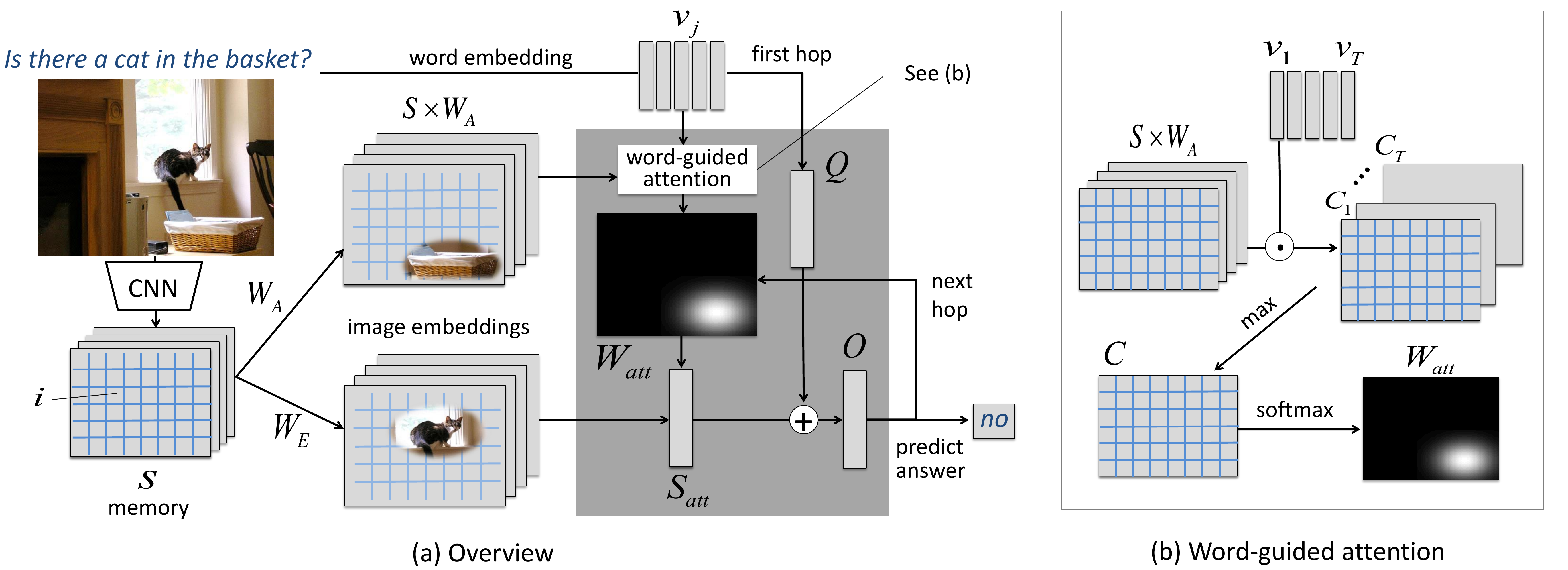}
\vspace{-0.05in}
\caption{Our proposed Spatial Memory Network for Visual Question Answering (SMem-VQA). (a) Overview. First, the CNN activation vectors $S=\{s_i\}$ at image locations $i$ are projected into the semantic space of the question word vectors $v_j$ using the ``attention'' visual embedding $W_A$ (Sec.~\ref{sec:smem}). The results are then used to infer spatial attention weights $W_{att}$ using the word-guided attention process shown in (b). 
(b) Word-guided attention. This process predicts attention determined by the question word that has the maximum correlation with embedded visual features at each location, e.g. choosing the word \textit{basket} to attend to the location of the basket in the above image (Sec.~\ref{sec:att1}).
The resulting spatial attention weights $W_{att}$ are then used to compute a weighted sum over the visual features embedded via a separate ``evidence'' transformation $W_E$, e.g., selecting evidence for the cat concept at the basket location. Finally, the weighted evidence vector $S_{att}$ is combined with the full question embedding $Q$ to predict the answer. An additional hop can repeat the process to gather more evidence (Sec.~\ref{sec:att3}). 
}
\label{fig:illus}
\vspace{-0.15in}
\end{figure*}

%%%%%%%%%%%%%%%%%%%%%%%%%%%%%%%%%%%%%%%%%%%%%%%%%%%%%%%%%%%%%%%%%%%%%%%%%%%%%%%%%%%%%%%%%%%%%
%\subsection{Overview}
We first give an overview of the proposed SMem-VQA network, illustrated in Fig.~\ref{fig:illus} (a). Sec.~\ref{sec:att1} details the word-guided spatial attention process of the first hop shown in Fig.~\ref{fig:illus} (b), and Sec.~\ref{sec:att3} describes adding a second hop into SMem-VQA network.

The input to our network is a question comprised of a variable-length sequence of words, and an image of fixed size.
Each word in the question is first represented as a one-hot vector in the size of the vocabulary, with a value of one only in the corresponding word position and zeros in the other positions. Each one-hot vector is then embedded into a real-valued word vector, $V=\{v_j \;|\;v_j\in \mathbb{R}^N; j=1,\cdots,T\}$, where $T$ is the maximum number of words in the question and $N$ is the dimensionality of the embedding space. Sentences with length less than $T$ are padded with special \textbf{$-1$} value, which are embedded to all-zero word vector.

The words in questions are used to compute attention over the visual memory, which contains extracted image features. The input image is processed by a convolutional neural network (CNN) to extract high-level $M$-dimensional visual features on a grid of spatial locations. 
Specifically, we use $S=\{s_i \;|\;s_i\in \mathbb{R}^M; i=1,\cdots,L\}$ to represent the spatial CNN features at each of the $L$ grid locations. In this paper, the spatial feature outputs of the last convolutional layer of GoogLeNet ($inception\_5b/output$) \cite{googlenet} are used as the visual features for the image. 

The convolutional image feature vectors at each location are  embedded into a common semantic space with the word vectors.
Two different embeddings are used: the ``attention'' embedding $W_A$ and the ``evidence'' embedding $W_E$. 
The attention embedding projects each visual feature vector such that its combination with the embedded question words generates the attention weight at that location. The evidence embedding detects the presence of semantic concepts or objects, and the embedding results are multiplied with attention weights and summed over all locations to generate the visual evidence vector $S_{att}$. 

Finally, the visual evidence vector is combined with the question representation and used to predict the answer for the given image and question.
In the next section, we describe the one-hop Spatial Memory network model and the specific attention mechanism it uses in more detail. 

%%%%%%%%%%%%%%%%%%%%%%%%%%%%%%%%%%%%%%%%%%%%%%%%%%%%%%%%%%%%%%%%%%%%%%%%%%%%%%%%%%%%%%%%%%%%%
\subsection{Word Guided Spatial Attention in One-Hop Model}\label{sec:att1}
Rather than using the bag-of-words question representation to guide attention,
the attention architecture in the first hop (Fig.~\ref{fig:illus}(b)) uses each word vector separately to extract correlated visual features in memory. 
The intuition is that the BOW representation may be too coarse, and letting each word select a related region may provide more fine-grained attention. 
The correlation matrix $C \in \mathbb{R}^{T\times L}$ between word vectors $V$ and visual features $S$ is computed as
\begin{equation}
C = V \cdot (S \cdot W_A + b_A)^T
\end{equation}
where $W_A \in \mathbb{R}^{M\times N}$ contains the attention embedding weights of visual features $S$, and $b_A \in \mathbb{R}^{L\times N}$ is the bias term. 
This correlation matrix is the dot product result of each word embedding and each spatial location's visual feature, thus each value in correlation matrix $C$ measures the similarity between each word and each location's visual feature. 

The spatial attention weights $W_{att}$ are calculated by taking maximum over the word dimension $T$ for the correlation matrix $C$, selecting the highest correlation value for each spatial location, and then applying the softmax function
\begin{equation}
W_{att} = \softmax(\max_{i=1,\cdots,T}(C_i)), ~C_i \in \mathbb{R}^L
\end{equation}
The resulting attention weights $W_{att} \in \mathbb{R}^{L}$ are high for selected locations and low for other locations, with the sum of weights equal to $1$. For instance, in the example shown in Fig.~\ref{fig:illus}, the question ``Is there a cat in the basket?'' produces high attention weights for the location of the basket because of the high correlation of the word vector for \textit{basket} with the visual features at that location. 

The evidence embedding $W_E$ projects visual features $S$ to produce high activations for certain semantic concepts. E.g., in Fig.~\ref{fig:illus}, it has high activations in the region containing the cat. The results of this evidence embedding are then multiplied by the generated attention weights $W_{att}$, and summed to produce the selected visual ``evidence'' vector $S_{att} \in \mathbb{R}^N$,
\begin{equation}
S_{att} = W_{att} \cdot (S \cdot W_E + b_E) \label{eqn:evidence}
\end{equation}
where $W_E \in \mathbb{R}^{M\times N}$ are the evidence embedding weights of the visual features $S$, and $b_E \in \mathbb{R}^{L\times N}$ is the bias term.
In our running example, this step accumulates \textit{cat} presence features at the \textit{basket} location. 

Finally, the sum of this evidence vector $S_{att}$ and the question embedding $Q$ is used to predict the answer for the given image and question.
For the question representation $Q$, we choose the bag-of-words (BOW). Other question representations, such as an LSTM, can also be used, however, BOW has fewer parameters yet has shown good performance. As noted in~\cite{shih2015look}, the simple BOW model performs roughly as well if not better than the sequence-based LSTM for the VQA task. Specifically, we compute
\begin{equation}
Q = W_Q \cdot V + b_Q
\end{equation}
where $W_Q \in \mathbb{R}^T$ represents the BOW weights for word vectors $V$, and $b_Q \in \mathbb{R}^{N}$ is the bias term. The final prediction $P$ is
\begin{equation}
P = \softmax(W_P \cdot f(S_{att} + Q) + b_P)\label{eqn:predict}
\end{equation}
where $W_P \in \mathbb{R}^{K\times N}$, bias term $b_P \in \mathbb{R}^{K}$, and $K$ represents the number of possible prediction answers. $f$ is the activation function, and we use ReLU here.
In our running example, this step adds the evidence gathered for \textit{cat} near the basket location to the question, and, since the cat was not found, predicts the answer ``no''. 
The attention and evidence computation steps can be optionally repeated in another hop, before predicting the final answer, as detailed in the next section. 

%%%%%%%%%%%%%%%%%%%%%%%%%%%%%%%%%%%%%%%%%%%%%%%%%%%%%%%%%%%%%%%%%%%%%%%%%%%%%%%%%%%%%%%%%%%%%
\subsection{Spatial Attention in Two-Hop Model}\label{sec:att3}
We can repeat hops to promote deeper inference, gathering additional evidence at each hop. Recall that the visual evidence vector $S_{att}$ is added to the question representation $Q$ in the first hop to produce an updated question vector,
\begin{equation}{O_{hop1} = S_{att} + Q}\end{equation}
On the next hop, this vector $O_{hop1} \in \mathbb{R}^{N}$ is used in place of the individual word vectors $V$ to extract additional correlated visual features to the whole question from memory and update the visual evidence.

The correlation matrix $C$ in the first hop provides fine-grained local evidence from each word vectors $V$ in the question, while the correlation vector $C_{hop2}$ in next hop considers the global evidence from the whole question representation $Q$. 
The correlation vector $C_{hop2} \in \mathbb{R}^L$ in the second hop is calculated by 
\begin{equation}
C_{hop2} = (S \cdot W_E + b_E) \cdot O_{hop1}
\end{equation}
where $W_E \in \mathbb{R}^{M\times N}$ should be the attention embedding weights of visual features $S$ in the second hop and $b_E \in \mathbb{R}^{L\times N}$ should be the bias term. Since the attention embedding weights in the second hop are shared with the evidence embedding in the first hop, so we directly use $W_E$ and $b_E$ from first hop here. 

The attention weights in the second hop $W_{att2}$ are obtained by applying the softmax function to the correlation vector $C_{hop2}$.
\begin{equation}
W_{att2} = \softmax(C_{hop2})
\end{equation}

Then, the correlated visual information in the second hop $S_{att2} \in \mathbb{R}^N$ is extracted using attention weights $W_{att2}$.
\begin{equation}
S_{att2} = W_{att2} \cdot (S \cdot W_{E_2} + b_{E_2})
\end{equation}
where $W_{E_2} \in \mathbb{R}^{M\times N}$ are the evidence embedding weights of visual features $S$ in the second hop, and $b_{E_2} \in \mathbb{R}^{L\times N}$ is the bias term. 

The final answer $P$ is predicted by combining the whole question representation $Q$, the local visual evidence $S_{att}$ from each word vector in the first hop and the global visual evidence $S_{att2}$ from the whole question in the second hop,
\begin{equation}
P = \softmax(W_P \cdot f(O_{hop1} + S_{att2}) + b_P)
\end{equation}
where $W_P \in \mathbb{R}^{K\times N}$, bias term $b_P \in \mathbb{R}^{K}$, and $K$ represents the number of possible prediction answers. $f$ is activation function. More hops can be added in this manner. 

The entire network is differentiable and is trained using stochastic gradient descent via standard backpropagation, allowing image feature extraction, image embedding, word embedding and answer prediction to be  jointly optimized on the training image/question/answer triples. 
\section{Experiments}
\vspace{-0.05in}

In this section, we conduct a series of experiments to evaluate our model. 
To explore whether the model learns to perform the spatial inference necessary for answering visual questions that explicitly require spatial reasoning, we design a set of experiments using synthetic visual question/answer data in Sec.~\ref{sec:synthetic1}. The experimental results of our model in standard datasets (DAQUAR~\cite{DBLP:journals/corr/MalinowskiF14} and VQA~\cite{DBLP:journals/corr/AntolALMBZP15} datasets) are reported in Sec.~\ref{sec:expstandard}.

%%%%%%%%%%%%%%%%%%%%%%%%%%%%%%%%%%%%%%%%%%%%%%%%%%%%%%%%%%%%%%%%%%%%%%%%%%%%%%%%%%%%%%%%%%%%%%%%%%%
\subsection{Exploring Attention on Synthetic Data}\label{sec:synthetic1}
%Several public datasets are available, however, 
The questions in the public VQA datasets are quite varied and difficult and often require common sense knowledge to answer (e.g., ``Does this man have 20/20 vision?'' about a person wearing glasses). Furthermore, past work~\cite{malinowski2015ask,DBLP:journals/corr/RenKZ15} showed that the question text alone (no image) is a very strong predictor of the answer.
Therefore, before evaluating on standard datasets, we would first like to
understand how the proposed model uses spatial attention to answer simple visual questions where the answer cannot be predicted from question alone. 
Our visualization demonstrates that the attention mechanism does learn to attend to objects and gather evidence via certain inference rules. 

%%%%%%%%%%% figure
\begin{figure*}[!t]
  \includegraphics[width=0.242\textwidth]{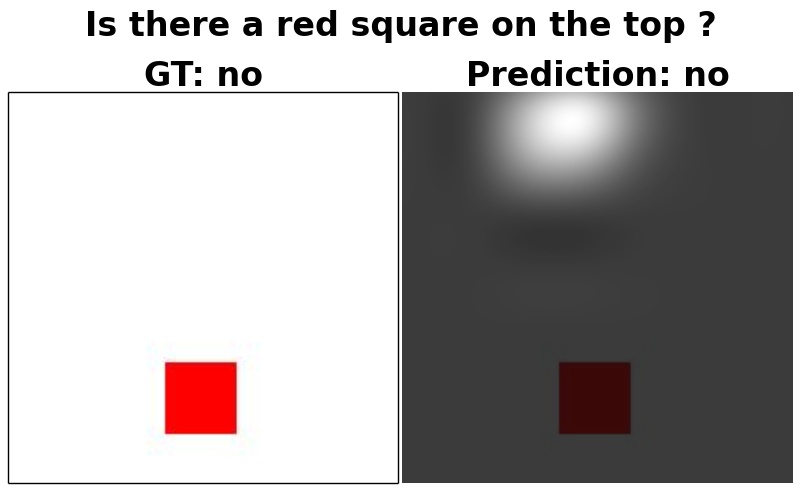}
  \includegraphics[width=0.242\textwidth]{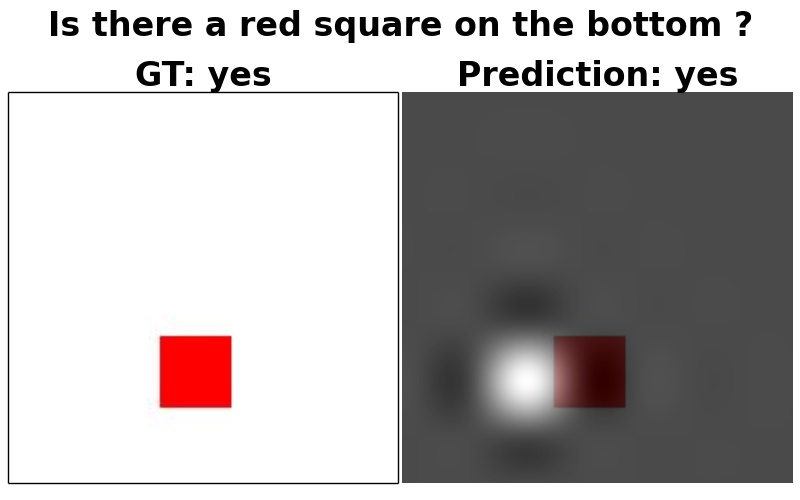}
  \includegraphics[width=0.242\textwidth]{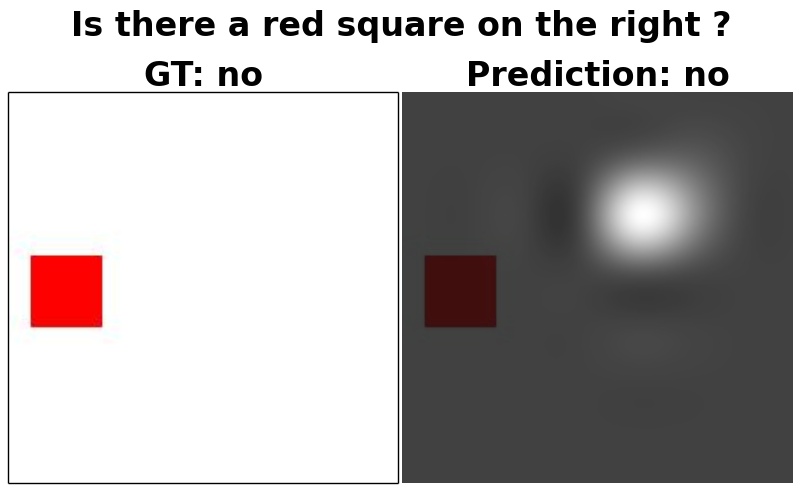}
  \includegraphics[width=0.242\textwidth]{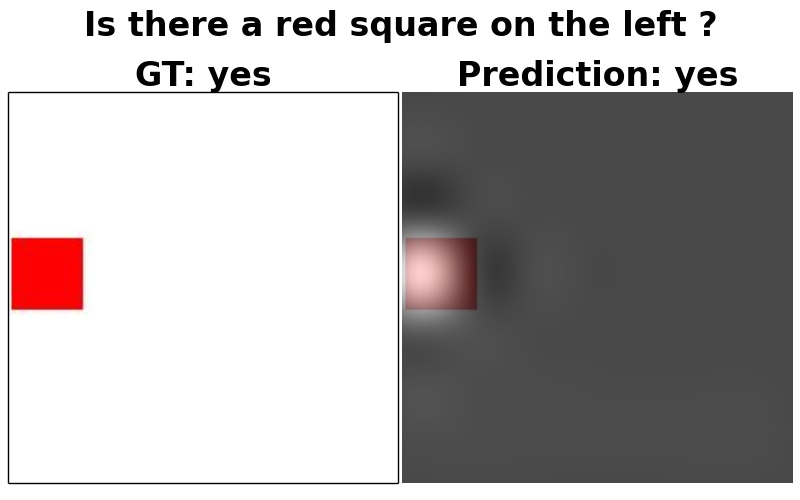}\\
  \includegraphics[width=0.242\textwidth]{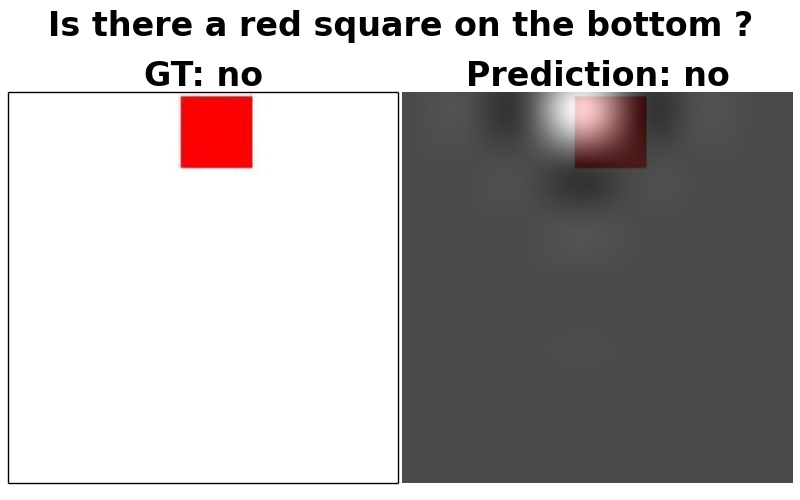}
  \includegraphics[width=0.242\textwidth]{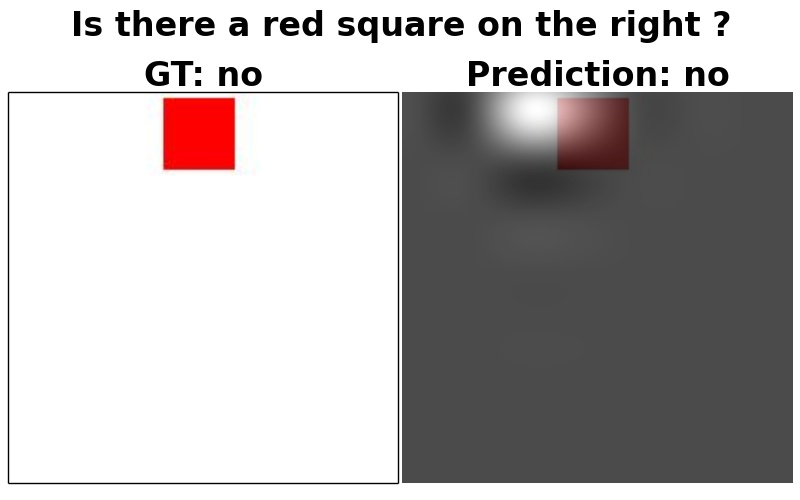}
  \includegraphics[width=0.242\textwidth]{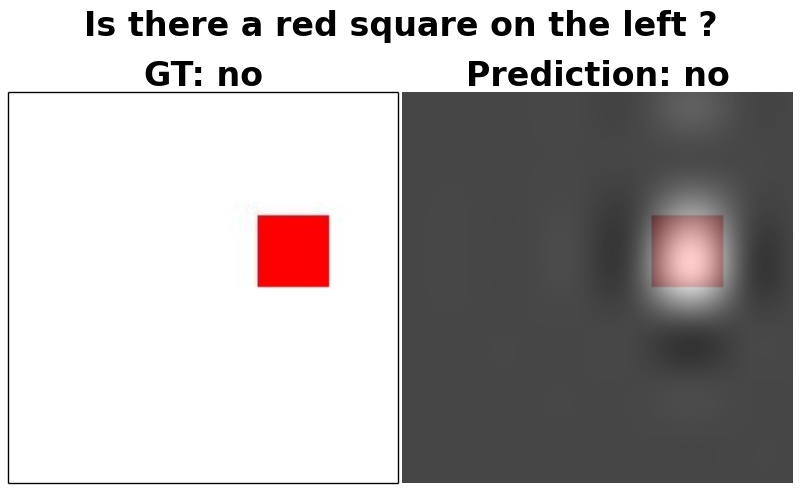}
  \includegraphics[width=0.242\textwidth]{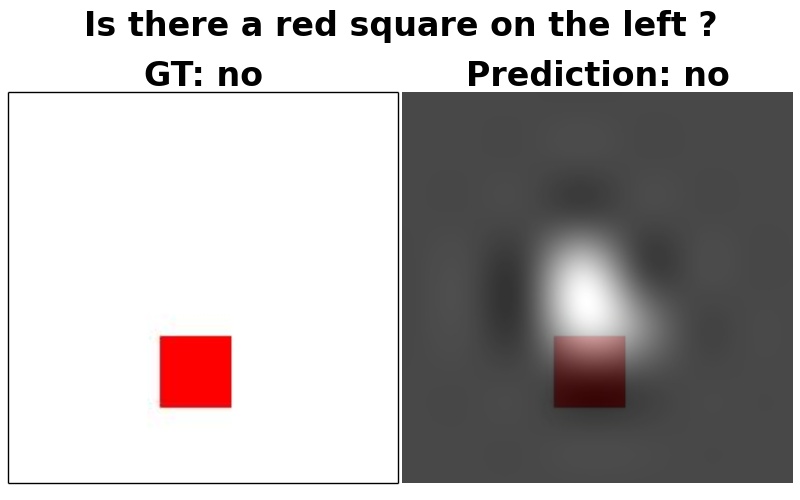}
\vspace{-0.1in}
\caption{\textbf{Absolute position experiment:} for each image and question pair, we show the original image (left) and the attention weights $W_{att}$ (right). 
The attention follows the following rules. The first rule (top row) looks at the position specified in question (top$\mid$bottom$\mid$right$\mid$left), if it contains a square, answer ``yes''; otherwise answer ``no''.
The second rule (bottom row) looks at the region where there is a square, and answers ``yes'' if the question contains that position and ``no'' for the other three positions.}\label{fig:red_square}
\vspace{-0.1in}
\end{figure*}

%%%%%%%%%%%%%%%%%%%%%%%%%%%%%%%%%%%%%%%%%%%%%%%%%%%%%%%%%%%%%%%%%%%%%%%%%%%%%%%%%%%%%%%%%%%%%%%%%%%
\vspace{-0.1in}
\subsubsection{Absolute Position Recognition}\label{sec:absolute}
We investigate whether the model has the ability to recognize the absolute location of the object in the image. We explore this by designing a simple task where an object (a red square) appears in some region of a white-background image, and the question is ``Is there a red square on the [top$\mid$bottom$\mid$left$\mid$right]?'' For each image, we randomly place the square in one of the four regions, and generate the four questions above, together with three ``no'' answers and one ``yes'' answer. The generated data is split into training and testing sets. 

Due to the simplicity of this synthetic dataset, the SMem-VQA one-hop model achieves 100\% test accuracy. However, the baseline model (iBOWIMG)~\cite{zhou2015simple} cannot infer the answer and only obtains accuracy of around 75\%, which is the prior probability of the answer ``no'' in the training set. The SMem-VQA one-hop model is equivalent to the iBOWIMG model if the attention weights in our one-hop model are set equally for each location, since the iBOWIMG model uses the mean pool of the convolutional feature ($inception\_5b/output$) in GoogLeNet that we use in SMem-VQA model. 
We check the visualization of the attention weights and find that the relationship between the high attention position and the answer can be expressed by logical expressions.
We show the attention weights of several typical examples in Fig.~\ref{fig:red_square} which reflect two logic rules:
1)~Look at the position specified in question (top$\mid$bottom$\mid$right$\mid$left), if it contains a square, then answer ``yes''; if it does not contain a square, then answer ``no''.
2)~Look at the region where there is a square, then answer ``yes'' for the question about that position and ``no'' for the questions about the other three positions.

In the iBOWIMG model, the mean-pooled GoogLeNet visual features lose spatial information and thus cannot distinguish images with a square in different positions. On the contrary, our SMem-VQA model can pay high attention to different regions according to the question, and generate an answer based on the selected region, using some learned inference rules.
This experiment demonstrates that the attention mechanism in our model is able to make absolute spatial location inference based on the spatial attention.

%%%%%%%%%%% figure
\begin{figure*}[t]
  \includegraphics[width=0.325\textwidth]{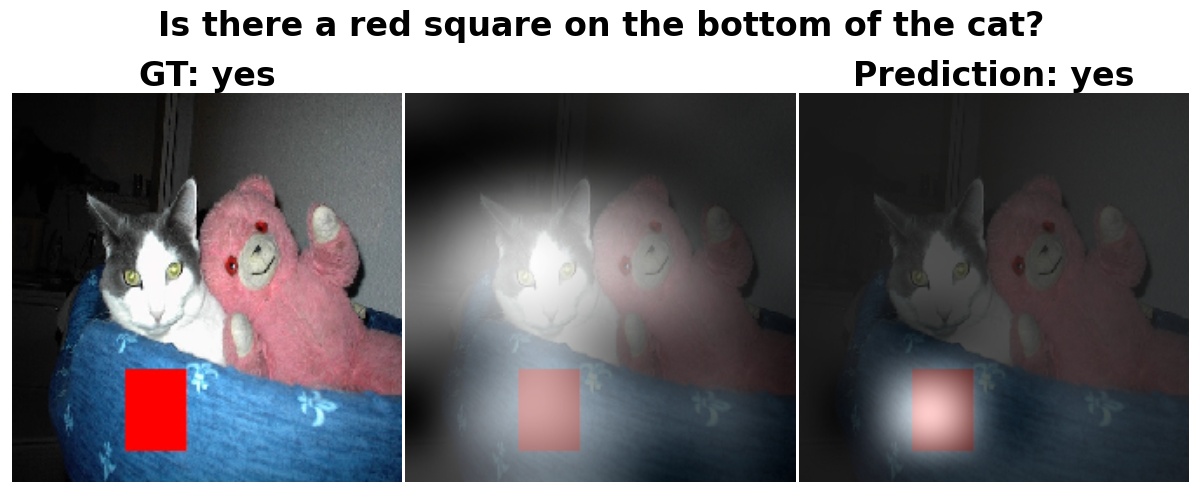}
  \includegraphics[width=0.325\textwidth]{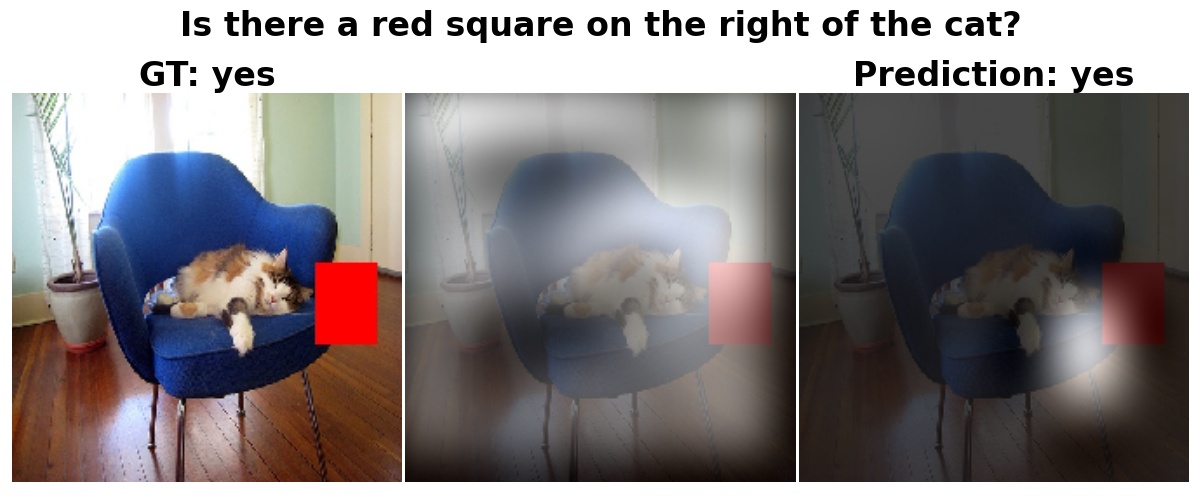}
  \includegraphics[width=0.325\textwidth]{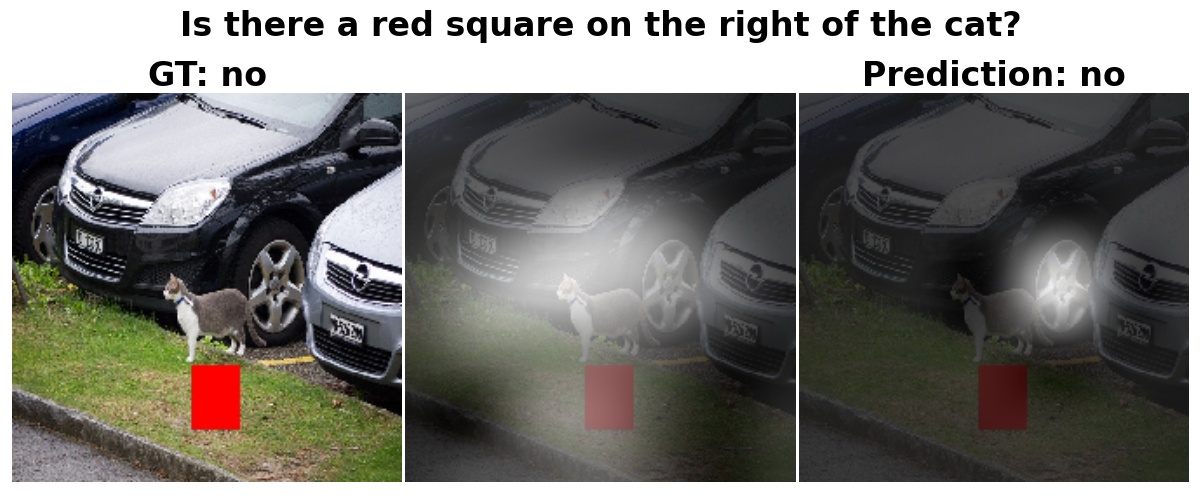}\\
  \includegraphics[width=0.325\textwidth]{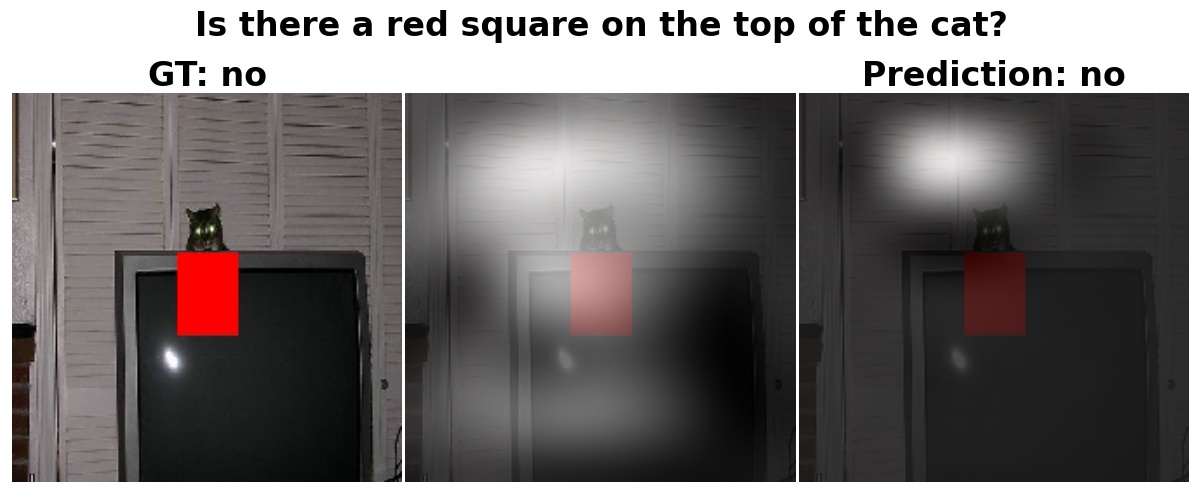}
  \includegraphics[width=0.325\textwidth]{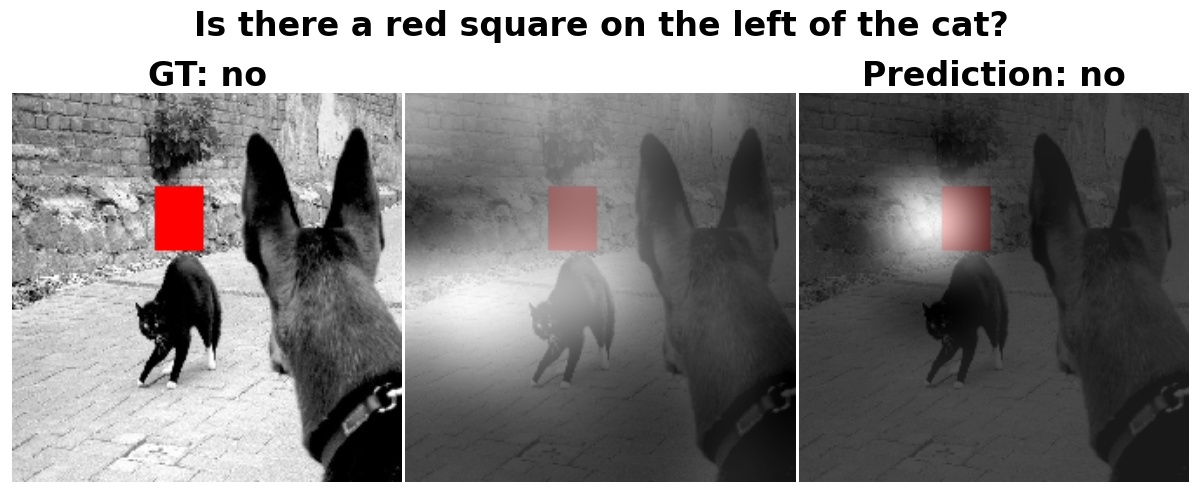}
  \includegraphics[width=0.325\textwidth]{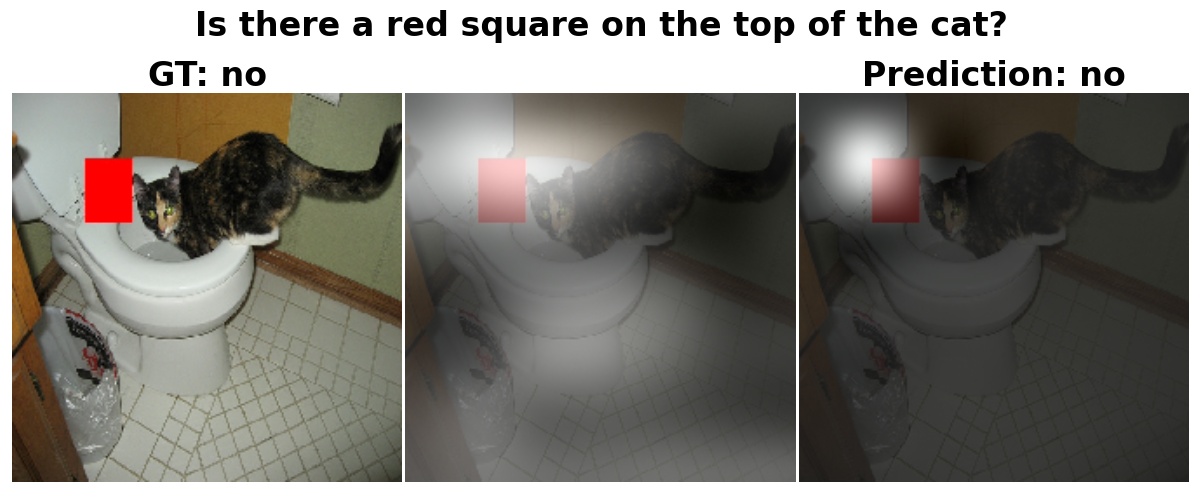}
\vspace{-0.05in}
\caption{\textbf{Relative position experiment:}
for each image and question pair, we show the original image (left), the evidence embedding $W_E$ of the convolutional layer (middle) and the attention weights $W_{att}$ (right). The evidence embedding $W_E$ has high activations on both cat and red square. 
The attention weights follow similar inference rules as in Fig.~\ref{fig:red_square}, with the difference that the attention position is around the cat.
%Rule#1: (top row) look at the specified position (top/bottom/right/left), if it contains a square, answer "yes"; otherwise answer "no".
%Rule#2: (bottom row) look at the region where there is a square, answer "yes" if the question contains that position and "no" if it contains one of the other three positions.
}\label{fig:cat_square}
\vspace{-0.2in}
\end{figure*}

%%%%%%%%%%%%%%%%%%%%%%%%%%%%%%%%%%%%%%%%%%%%%%%%%%%%%%%%%%%%%%%%%%%%%%%%%%%%%%%%%%%%%%%%%%%%%%%%%%%
\vspace{-0.1in}
\subsubsection{Relative Position Recognition}
In order to check whether the model has the ability to infer the  position of one object \textit{relative} to another object,
we collect all the cat images from the MS COCO Detection dataset~\cite{lin2014microsoft}, and add a red square on the [top$\mid$bottom$\mid$left$\mid$right] of the bounding box of the cat in the images.
For each generated image, we create four questions, ``Is there a red square on the [top$\mid$bottom$\mid$left$\mid$right] of the cat?'' together with three ``no'' answers and one ``yes'' answer. 
We select 2639 training cat images and 1395 testing cat images from MS COCO Detection dataset. 

Our SMem-VQA one-hop model achieves 96\% test accuracy on this synthetic task, while the baseline model (iBOWIMG) accuracy is around 75\%.
We also check that another simple baseline that predicts the answer based on the absolute position of the square in the image gets around 70\% accuracy. 
We visualize the evidence embedding $W_E$ features and the attention weights $W_{att}$ of several typical examples in Fig.~\ref{fig:cat_square}.
The evidence embedding $W_E$ has high activations on the cat and the red square, while the attention weights pay high attention to certain locations around the cat.
We can analyze the attention in the correctly predicted examples using the same rules as in absolute position recognition experiment. 
These rules still work, but the position is relative to the cat object:
1)~Check the specified position relative to the cat, if it finds the square, then answer ``yes'', otherwise ``no''; 2)~Find the square, then answer ``yes'' for the specified position, and answer ``no'' for the other positions around the cat.
We also check the images where our model makes mistakes, and find that the mistakes mainly occur in images with more than one cats. The red square appears near only one of the cats in the image, but our model might make mistakes by focusing on the other cats.
We conclude that our SMem-VQA model can infer the relative spatial position based on the spatial attention around the specified object, which can also be represented by some logical inference rules.

%%%%%%%%%%%%%%%%%%%%%%%%%%%%%%%%%%%%%%%%%%%%%%%%%%%%%%%%%%%%%%%%%%%%%%%%%%%%%%%%%%%%%%%%%%%%%%%%%%%
%%%%%%%%%%%%%% table
\begin{table}[!t]
\centering
\caption{Accuracy results on the DAQUAR dataset (in percentage).}
\small
 \begin{tabular}{l || c c c} 
 \hline
 ~ & DAQUAR\\ \hline
 Multi-World~\cite{DBLP:journals/corr/MalinowskiF14} & 12.73 \\ %\hline
 Neural-Image-QA~\cite{malinowski2015ask} & 29.27  \\ %\hline
 Question LSTM~\cite{malinowski2015ask} & 32.32 \\ %\hline
 VIS+LSTM~\cite{DBLP:journals/corr/RenKZ15} & 34.41 \\ %\hline
 Question BOW~\cite{DBLP:journals/corr/RenKZ15} & 32.67 \\ %\hline
 IMG+BOW~\cite{DBLP:journals/corr/RenKZ15} & 34.17 \\ \hline
 %2-VIS+BLSTM ~\cite{DBLP:journals/corr/RenKZ15} &  - & 35.78  & -\\ \hline \hline
 %Question One-Hop & 53.37 & 36.03 & - \\ %\hline   
 SMem-VQA One-Hop & 36.03 \\ %\hline   
 SMem-VQA Two-Hop & \bf{40.07} \\ \hline 
 %one dimension convolution hierarchy model\cite{ma2015learning} &  - & 0.3933 \cite{ma2015learning} & - \\ \hline
 \end{tabular}
\label{fig:baseline}
\vspace{-0.2in}
\end{table}

\subsection{Experiments on Standard Datasets}\label{sec:expstandard}
%\vspace{-0.08in}
\subsubsection{Results on DAQUAR}
The DAQUAR dataset is a relatively small dataset which builds on the NYU Depth Dataset V2~\cite{Silberman:ECCV12}. We use the reduced DAQUAR dataset~\cite{DBLP:journals/corr/MalinowskiF14}. The evaluation metric for this dataset is 0-1 accuracy. 
The embedding dimension is 512 for our models running on the DAQUAR dataset. 
We use several reported models on DAQUAR as baselines, which are listed below:\\
%\noindent
{$\bullet$} {\bf{Multi-World}}~\cite{DBLP:journals/corr/MalinowskiF14}: an approach based on handcrafted features using a semantic parse of the question and scene analysis of the image combined in a latent-world Bayesian framework.\\ 
{$\bullet$} {\bf{Neural-Image-QA}}~\cite{malinowski2015ask}: uses an LSTM to encode the question and then decode the hidden information into the answer. The image CNN feature vector is shown at each time step of the encoding phase.\\
{$\bullet$} {\bf{Question LSTM}}~\cite{malinowski2015ask}: only shows the question to the LSTM to predict the answer without any image information.\\
{$\bullet$} {\bf{VIS+LSTM}}~\cite{DBLP:journals/corr/RenKZ15}: similar to Neural-Image-QA, but only shows the image features to the LSTM at the first time step, and the question in the remaining time steps to predict the answer.\\
{$\bullet$} {\bf{Question BOW}}~\cite{DBLP:journals/corr/RenKZ15}: only uses the BOW question representation and a single hidden layer neural network to predict the answer, without any image features.\\
{$\bullet$} {\bf{IMG+BOW}}~\cite{DBLP:journals/corr/RenKZ15}: concatenates the BOW question representation with image features, and then uses a single hidden layer neural network to predict the answer. This model is similar to the iBOWIMG baseline model in~\cite{zhou2015simple}.

\begin{figure*}[t]
  \includegraphics[width=0.5\textwidth]{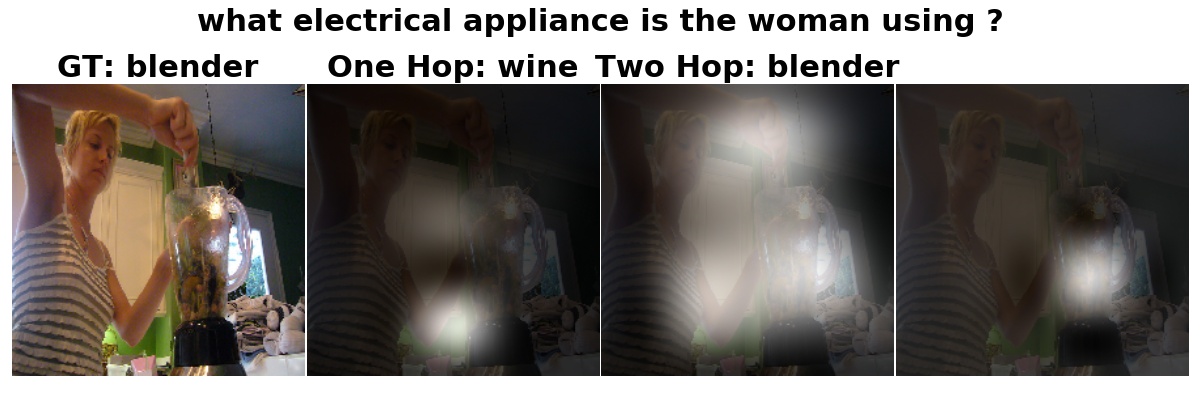}
  \includegraphics[width=0.5\textwidth]{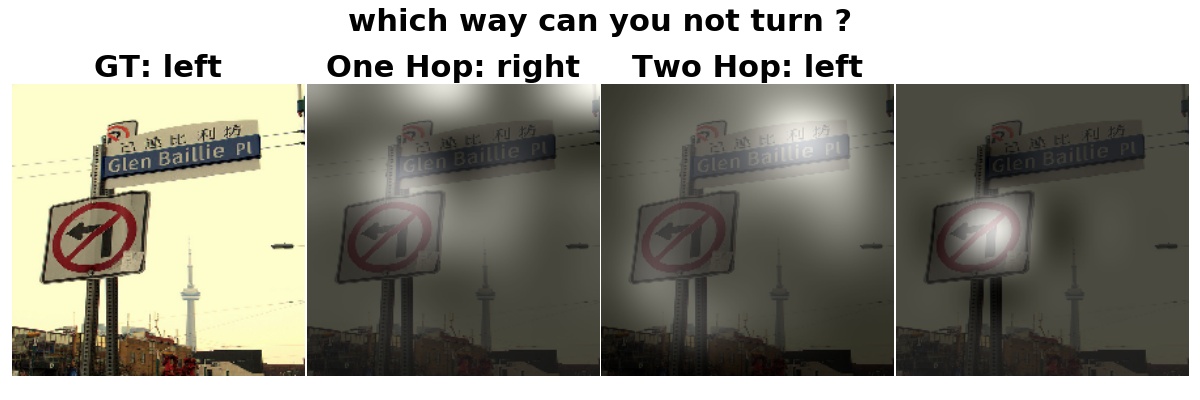}\\
  \includegraphics[width=0.5\textwidth]{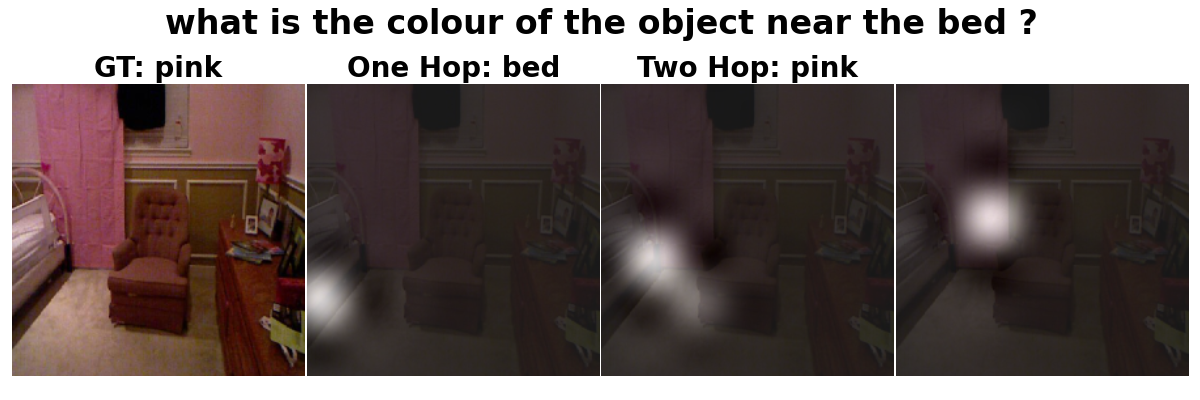}
  \includegraphics[width=0.5\textwidth]{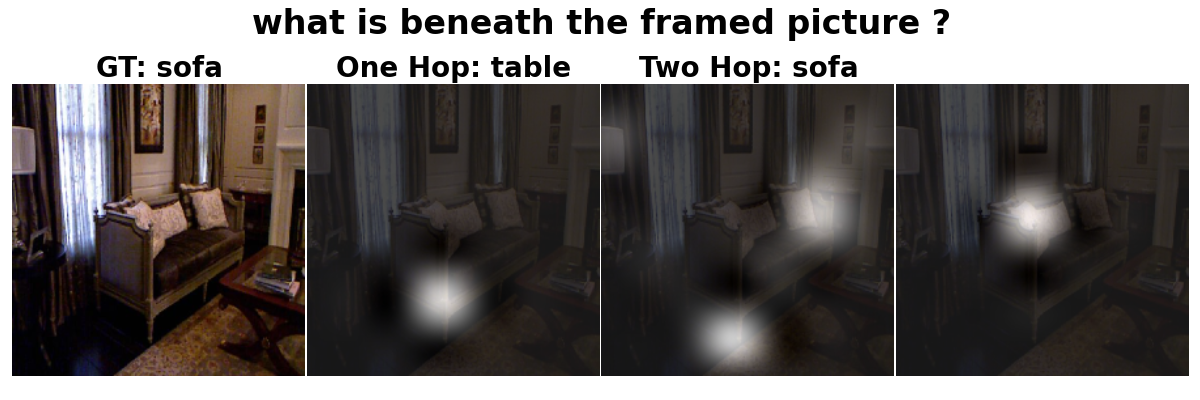}
\vspace{-0.25in}
\caption{Visualization of the spatial attention weights in the SMem-VQA One-Hop and Two-Hop models on VQA (top row) and DAQUAR (bottom row) datasets. For each image and question pair, we show the original image, the attention weights $W_{att}$ of the One-Hop model, and the two attention weights $W_{att}$ and $W_{att2}$ of the Two-Hop model in order.}\label{fig:2hopVQA}
\vspace{-0.15in}
\end{figure*}

Results of our SMem-VQA model on the DAQUAR dataset and the baseline model results reported in previous work are shown in Tab.~\ref{fig:baseline}. 
From the DAQUAR result in Tab.~\ref{fig:baseline}, we see that models based on deep features significantly outperform the Multi-World approach based on hand-crafted features. Modeling the question only with either the LSTM model or Question BOW model does equally well in comparison, indicating the the question text contains important prior information for predicting the answer. Also, on this dataset, the VIS+LSTM model achieves better accuracy than Neural-Image-QA model; the former shows the image only at the first timestep of the LSTM, while the latter does so at each timestep. In comparison, both our One-Hop model and Two-Hop spatial attention models outperform the IMG+BOW, as well as the other baseline models.
A major advantage of our model is the ability to visualize the inference process in the deep network. To illustrate this, two attention weights visualization examples in SMem-VQA One-Hop and Two-Hop models on DAQUAR dataset are shown in Fig.~\ref{fig:2hopVQA} (bottom row).

%%%%%%%%%%% figure
% \vspace{-0.1in}
\subsubsection{Results on VQA}

The VQA dataset is a recent large dataset based on MS COCO~\cite{lin2014microsoft}. We use the full release (V1.0) open-ended dataset, which contains a train set and a val set. Following standard practice, we choose the top 1000 answers in train and val sets as possible prediction answers, and only keep the examples whose answers belong to these 1000 answers as training data. The question vocabulary size is 7477 with the word frequency of at least three.
Because of the larger training size, the embedding dimension is 1000 on the VQA dataset.
We report the test-dev and test-standard results from the VQA evaluation server.
The server evaluation uses the evaluation metric introduced by~\cite{DBLP:journals/corr/AntolALMBZP15}, which gives partial credit to certain synonym answers:
%\begin{equation}
%Acc({\color{red}ans}) = \min\left\{ \frac{\text{\# human that said }{\color{red} ans}}{3},1\right\}
%$Acc({ans}) = \min\left\{ \frac{\text{\# humans that said }{ans}}{3},1\right\}$.
$Acc({ans}) = \min\left\{ (\text{\# humans that said }{ans})/3,1\right\}$.
%\end{equation}

For the attention models, we do not mirror the input image when using the CNN to extract convolutional features, since this might cause confusion about the spatial locations of objects in the input image.
%The input image size for CNN is $224 \times 224$.
The optimization algorithm used is stochastic gradient descent (SGD) with a minibatch of size 50 and momentum of 0.9.
The base learning rate is set to be 0.01 which is halved every six epoches. Regularization, dropout and L2 norm are cross-validated and used.

%%%%%%%%%%%%%% another candidate table with test-standard
\begin{table}[!t]
\centering
\caption{Test-dev and test-standard results on the Open-Ended VQA dataset (in percentage). Models with ${}^\ast$ use external training data in addition to the VQA dataset.}
\scriptsize
 \begin{tabular}{l || c c c c || c c c c} 
 \hline
 ~ & \multicolumn{4}{c||}{test-dev}  & \multicolumn{4}{c}{test-standard}\\ 
 ~ & \bf{Overall}  & yes/no  & number  & others & \bf{Overall}  & yes/no  & number  & others\\ \hline
 LSTM Q+I~\cite{DBLP:journals/corr/AntolALMBZP15} & 53.74 & 78.94  & 35.24  & 36.42 & 54.06 & -  & -  & -\\ %\hline
 ACK${}^\ast$~\cite{wu2015ask} & 55.72 & 79.23  & 36.13  & 40.08 & 55.98 & 79.05  & 36.10  & 40.61\\ %\hline
 DPPnet${}^\ast$~\cite{noh2015image} & 57.22 & 80.71  & 37.24  & 41.69 & 57.36 & 80.28  & 36.92  & 42.24\\ %\hline
 iBOWIMG~\cite{zhou2015simple} & 55.72 & 76.55  & 35.03  & 42.62 & 55.89 & 76.76  & 34.98  & 42.62\\ \hline 
 SMem-VQA One-Hop & 56.56 & 78.98 & 35.93  & 42.09 & - & -  & -  & -\\ %\hline   
 SMem-VQA Two-Hop & \textbf{57.99} & \textbf{80.87}  & \textbf{37.32}  & \textbf{43.12} & \textbf{58.24} & \textbf{80.8}  & \textbf{37.53}  & \textbf{43.48}\\ \hline 
 \end{tabular}
\label{fig:baseline2}
\vspace{-0.15in}
\end{table}

For the VQA dataset, we use the simple iBOWIMG model in~\cite{zhou2015simple} as one baseline model, which beats most existing VQA models currently on arxiv.org. We also compare to two models in~\cite{wu2015ask}\cite{noh2015image} which have comparable or better results to the iBOWIMG model. These three baseline models as well the best model in VQA dataset paper~\cite{DBLP:journals/corr/AntolALMBZP15} are listed in the following:\\
%\noindent
{$\bullet$} {\bf{LSTM Q+I}}~\cite{DBLP:journals/corr/AntolALMBZP15}: uses the element-wise multiplication of the LSTM encoding of the question and the image feature vector to predict the answer. This is the best model in the VQA dataset paper.\\
{$\bullet$} {\bf{ACK}}~\cite{wu2015ask}: shows the image attribute features, the generated image caption and relevant external knowledge from knowledge base to the LSTM at the first time step, and the question in the remaining time steps to predict the answer.\\
{$\bullet$} {\bf{DPPnet}}~\cite{noh2015image}: uses the Gated Recurrent Unit (GRU) representation of question to predict certain parameters for a CNN classification network. They pre-train the GRU for question representation on a large-scale text corpus to improve the GRU generalization performance.\\
{$\bullet$} {\bf{iBOWIMG}}~\cite{zhou2015simple}: concatenates the BOW question representation with image feature (GoogLeNet), and uses a softmax classification to predict the answer. 

The overall accuracy and per-answer category accuracy for our SMem-VQA models and the four baseline models on VQA dataset are shown in Tab.~\ref{fig:baseline2}. From the table, we can see that the SMem-VQA One-Hop model obtains slightly better results compared to the iBOWIMG model. However, the SMem-VQA Two-Hop model achieves an improvement of 2.27\% on test-dev and 2.35\% on test-standard compared to the iBOWIMG model, demonstrating the value of spatial attention. The SMem-VQA Two-Hop model also shows best performance in the per-answer category accuracy. 
The SMem-VQA Two-Hop model has slightly better result than the DPPnet model. 
The DPPnet model uses a large-scale text corpus to pre-train the Gated Recurrent Unit (GRU) network for question representation.
Similar pre-training work on extra data to improve model accuracy has been done in~\cite{venugopalan2014translating}.
Considering the fact that our model does not use extra data to pre-train the word embeddings, its results are very competitive.
We also experiment with adding a third hop into our model on the VQA dataset, but the result does not improve further.

\begin{figure*}[t]
  \includegraphics[width=0.5\textwidth]{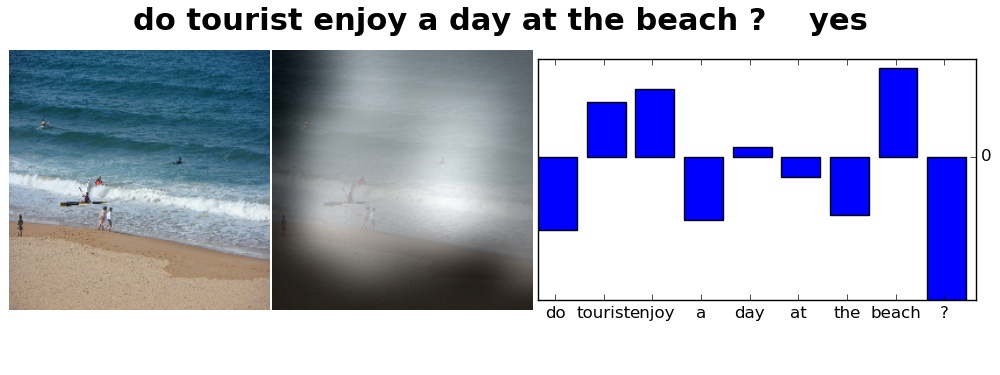}
  \includegraphics[width=0.5\textwidth]{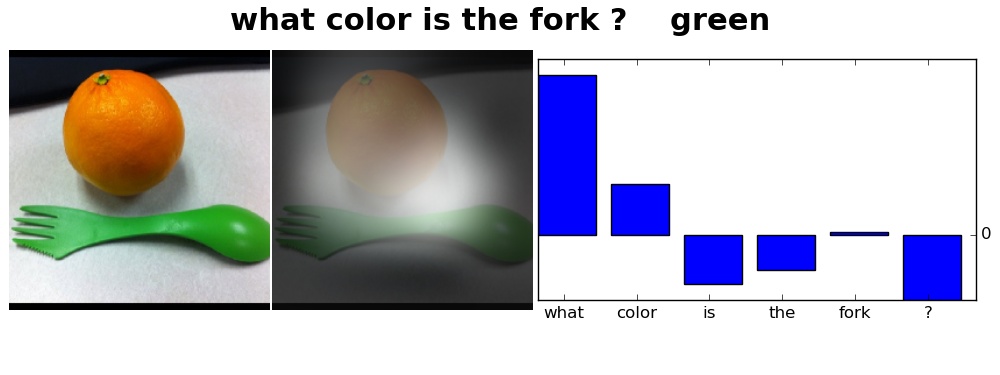}\\
  \includegraphics[width=0.5\textwidth]{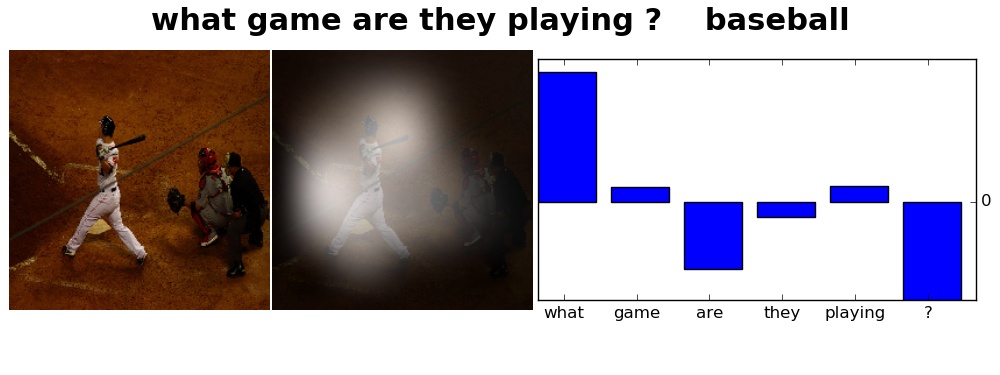}
  \includegraphics[width=0.5\textwidth]{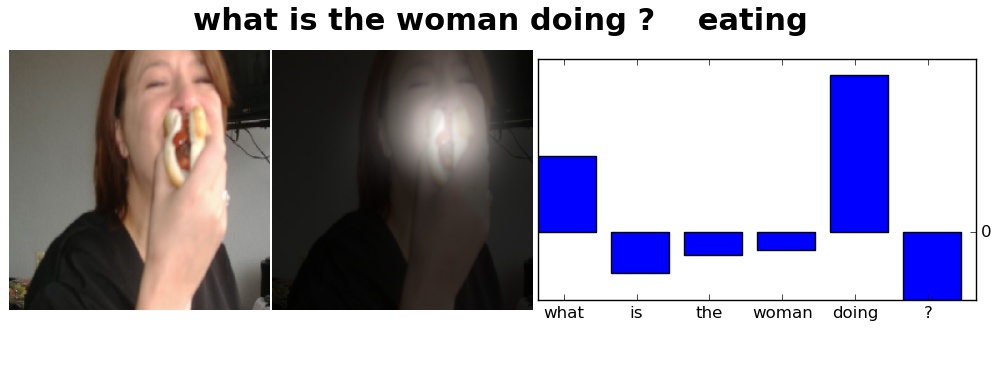}
\vspace{-0.3in}
\caption{
Visualization of the original image (left), the spatial attention weights $W_{att}$ in the first hop (middle) and one correlation vector from the correlation matrix $C$ for the location with highest attention weight in the SMem-VQA Two-Hop model on the VQA dataset.
Higher values in the correlation  vector indicate  stronger correlation of that word with the chosen location's image features.}
\label{fig:VQA_hop1Atten_wordAtten}
\vspace{-0.15in}
\end{figure*}

The attention weights visualization examples for the SMem-VQA One-Hop and Two-Hop models on the VQA dataset are shown in Fig.~\ref{fig:2hopVQA} (top row). From the visualization, we can see that the two-hop model collects supplementary evidence for inferring the answer, which may be necessary to achieve an improvement on these complicated real-world datasets. We also visualize the fine-grained alignment in the first hop of our SMem-VQA Two-Hop model in Fig.~\ref{fig:VQA_hop1Atten_wordAtten}. 
%through visualizing the attention weights and the correlation value vector from the correlation matrix $C$ for the location with highest attention weight
The correlation vector values (blue bars) measure the correlation between image regions and each word vector in the question. Higher values indicate stronger correlation of that particular word with the specific location's image features. We observe that the fine-grained visual evidence collected using each local word vector, together with the global visual evidence from the whole question, complement each other to infer the correct answer for the given image and question, as shown in Fig.~\ref{fig:concept}.

%%%%%%%%%%% figure

\vspace{-0.1in}
\section{Conclusion}
\vspace{-0.1in}
In this paper, we proposed the Spatial Memory Network for VQA, a memory network architecture with a spatial attention mechanism adapted to the visual question answering task. We proposed a set of synthetic spatial questions and demonstrated that our model learns inference rules based on spatial attention through attention weight visualization. Evaluation on the challenging DAQUAR and VQA datasets showed  improved results over previously published models. Our model can be used to visualize the inference steps learned by the deep network, giving some insight into its processing. Future work may include further exploring the inference ability of our SMem-VQA model  and exploring other VQA attention models.

%\clearpage

\bibliographystyle{splncs}
\bibliography{egbib}
\end{document}